%% file: footgait3D.tex
\documentclass{article}

\usepackage{arxiv}

 \usepackage{textgreek}      
 \usepackage{xcolor}         
 \usepackage{multirow} 
 \usepackage{amsmath}
 \usepackage{enumitem}
 \usepackage{makecell}

\usepackage[utf8]{inputenc} 
\usepackage[T1]{fontenc}    
\usepackage{hyperref}       
\usepackage{url}            
\usepackage{booktabs}       
\usepackage{amsfonts}       
\usepackage{nicefrac}       
\usepackage{microtype}      
\usepackage{lipsum}		
\usepackage{graphicx}
\usepackage{natbib}
\usepackage{doi}

\renewcommand{\cite}[1]{\citep{#1}}

\title{FootGait3D: A Multi-View High-Resolution Foot-Ankle Complex Point Cloud Dataset During Gait for Occlusion-Robust 3D Completion}

\author{ {Jie-Wen Li$^{1,*}$, Zi-Han Ye$^{2,*}$, Qingyuan Zhou$^{2}$, Jiayi Song$^{2}$, Ying He$^{4}$, Ben Fei$^{3,\dagger}$, Wen-Ming Chen$^{1,5,\dagger}$} \\
	$^1$ College of Biomedical Engineering, Fudan University, \\ $^2$ College of Computer Science and Artificial Intelligence, Fudan University, \\ $^3$ Department of Information Engineering, The Chinese University of Hong Kong, \\ $^4$ School of Computer Science and Engineering, Nanyang Technological University\\$^5$ Shanghai Innovation Institute \\
	\texttt{ljw285@163.com, 22300240027@m.fudan.edu.cn, benfei@cuhk.edu.hk, chenwm@fudan.edu.cn}}

\date{}

\begin{document}

\maketitle

\input{Sections/Abstract}

\newcommand\blfootnote[1]{%
\begingroup
\renewcommand\thefootnote{}\footnote{#1}%
\addtocounter{footnote}{-1}%
\endgroup
}
\blfootnote{{$*$}Equal contribution, {$\dagger$}Corresponding author.}


\input{Sections/Introduction}

\input{Sections/Related_Work}

\input{Sections/Dataset_Description}

\input{Sections/Benchmark_and_Evaluations}

\input{Sections/Discussion}

\bibliographystyle{unsrtnat}  
\bibliography{references}

\newpage
\input{appendix}

\end{document}

%% file: Sections/Abstract.tex
\begin{abstract}
The kinematics analysis of foot-ankle complex during gait is essential for advancing biomechanical research and clinical assessment. 
Collecting accurate surface geometry data from the foot and ankle during dynamic gait conditions is inherently challenging due to swing foot occlusions and viewing limitations.
Thus, this paper introduces FootGait3D, a novel multi-view dataset of high-resolution ankle-foot surface point clouds captured during natural gait. 
Different from existing gait datasets that typically target whole-body or lower-limb motion, FootGait3D focuses specifically on the detailed modeling of the ankle-foot region, offering a finer granularity of motion data.  
To address this, FootGait3D consists of 8,403 point cloud frames collected from 46 subjects using a custom five-camera depth sensing system. 
Each frame includes a complete 5-view reconstruction of the foot and ankle (serving as ground truth) along with partial point clouds obtained from only four, three, or two views. 
This structured variation enables rigorous evaluation of 3D point cloud completion methods under varying occlusion levels and viewpoints. 
Our dataset is designed for shape completion tasks, facilitating the benchmarking of state-of-the-art single-modal (e.g., PointTr, SnowflakeNet, Anchorformer) and multi-modal (e.g., SVDFormer, PointSea, CSDN) completion networks on the challenge of recovering the full foot geometry from occluded inputs.
FootGait3D has significant potential to advance research in biomechanics and multi-segment foot modeling, offering a valuable testbed for clinical gait analysis, prosthetic design, and robotics applications requiring detailed 3D models of the foot during motion. The dataset is now available at \href{https://huggingface.co/datasets/ljw285/FootGait3D}{https://huggingface.co/datasets/ljw285/FootGait3D}. 
\end{abstract}

%% file: Sections/Introduction.tex
\section{Introduction}
\label{sec:introduction}

Gait analysis is fundamental to understanding human locomotion, with applications spanning clinical diagnosis, rehabilitation engineering, sports biomechanics, and prosthetics design~\cite{buldt2013relationship,hulleck2022present,baker2006gait,buschmann2009humanoid,sethi2022comprehensive}. 
Within this domain, characterizing the intricate kinematics of the foot and ankle complex is crucial, as it plays a vital role in shock absorption, stability, and propulsion during walking~\cite{barnes2008association}. 
Traditional gait analysis is highly dependent on marker-based motion capture systems~\cite{carson2001kinematic}. 
While providing valuable kinematic data, these systems suffer from limitations such as skin motion artifacts, laborious marker placement, and the inability to capture detailed surface geometry~\cite{wade2022applications}. 
To address this limitation, recent advances in markerless motion capture technologies, particularly those that utilize depth sensors and point cloud data, offer promising alternatives~\cite{mentiplay2015gait,benedek2016lidar,vafadar2021novel}. 
Point clouds provide dense three-dimensional surface information, allowing a more comprehensive representation of anatomical structures compared to sparse marker sets~\cite{xiao2023unsupervised}. 
This rich geometric data is indispensable not only in various fields such as robotics~\cite{zhu2024point}, autonomous driving~\cite{cui2021deep}, and medical image analysis~\cite{elyan2022computer}, but also holds great potential for detailed foot kinematic analysis and the development of subject-specific biomechanical models. 

However, obtaining complete point clouds of the foot-ankle complex during real-world gait poses inherent challenges~\cite{shultz2011quantifying,clark2019three}. 
The main reasons for incompleteness include occlusion of the stance foot by the contralateral swing foot and the inherent difficulty in capturing the plantar surface using externally placed cameras~\cite{van2014dynamic},which frequently leads to missing data. While point cloud completion techniques aim to reconstruct complete shapes from such partial data, their development and evaluation under realistic conditions remain significantly constrained. This is primarily because most real-world gait datasets~\cite{zheng2022gait,boyne2022find,boyne2024found} lack both the ground-truth complete shapes of the foot–ankle complex and the paired partial–complete samples, both of which are critical for supervised learning and accurately assessing completion performance on authentic data.

The scarcity of real paired data has led most point cloud completion methods to rely primarily on large-scale synthetic datasets for training, such as PCN~\cite{yuan2018pcn} or ShapeNet~\cite{yu2021pointr}, which consist of synthetic CAD models. 
The common practice involves training or fine-tuning on these synthetic datasets and then evaluating the model's generalization capability by testing directly on partial real-world data (e.g., evaluating object completion models on scans from datasets like KITTI~\cite{geiger2012we}). 
However, due to the absence of ground-truth completion data in the KITTI dataset, models struggle to learn real-world geometry and tend to mispredict details under unseen noise and occlusion patterns~\cite{mirza2021robustness,zendel2015cv}. 
The gap between clean, synthetic data and noisy, non-uniformly sparse real-world data limits the generalization performance of these models. 
Consequently, while models may perform well on synthetic benchmarks, they frequently struggle to produce accurate and robust completions for real-world scenarios.

To bridge this critical gap and foster the development of more practical point cloud completion algorithms, we introduce FootGait3D. This novel dataset provides paired partial and complete real-world point clouds of the human foot-ankle complex captured dynamically during gait using a markerless multi-depth-camera system. FootGait3D provides a unique resource for biomechanics research requiring detailed, dynamic foot geometry. Moreover, it establishes a challenging benchmark for the training and evaluation of point-cloud completion algorithms directly on real-world dynamic data. By offering diverse degrees of realistic incompleteness paired with ground truth, it enables a comprehensive assessment of existing methods under authentic conditions. 
In summary, the contributions of this work are summarized as follows:

\begin{itemize}[leftmargin=1em, itemsep=0pt, topsep=0pt, parsep=0pt]
    \item The introduction of FootGait3D, a large-scale, real-world point cloud dataset of the foot-ankle complex during gait, captured via a markerless system based on multi-depth cameras.
    \item The provision of paired complete and partial point clouds, specifically designed to serve as a benchmark for point-cloud completion algorithms operating on dynamic, biological data.
    \item Initial benchmark results for state-of-the-art point-cloud completion methods, highlighting the challenges posed by this real-world dataset.
\end{itemize}

%% file: Sections/Related_Work.tex
\section{Related Work}
\label{sec:relatedwork}

\textbf{Point Cloud Datasets for Completion. } 
Data-driven point-cloud completion relies heavily on large-scale datasets.  Much of the foundational work has relied on synthetic datasets derived from 3D CAD models. The PCN dataset~\cite{yuan2018pcn}, generated from ShapeNet~\cite{chang2015shapenet}, provides paired partial and complete point clouds across 8 object categories and has been widely adopted. Similarly, ShapeNet55 and ShapeNet34~\cite{yu2021pointr}, also derived from the ShapeNet repository, offer a larger diversity of object categories (55 and 34, respectively) and are frequently used for evaluating generalization capabilities. While instrumental in driving algorithmic development, these datasets primarily consist of clean, uniformly sampled point clouds of rigid objects, lacking the noise, non-uniform density, dynamic deformation, and potential topological complexity often encountered in real-world captures.

Several real-world datasets exist, capturing diverse scenes and objects. KITTI~\cite{geiger2012we} provides LiDAR point clouds from autonomous driving scenarios, featuring vehicles, pedestrians, and cyclists in outdoor environments. ScanNet~\cite{dai2017scannet} offers richly annotated 3D reconstructions of indoor scenes, captured using RGB-D sensors. While these datasets offer valuable real-world geometric data, they often lack the paired partial-and-complete ground truth structures specifically designed for supervised point cloud completion training. Furthermore, they do not focus on dynamic, deformable biological structures like the human foot during gait. Datasets like RealPC~\cite{pathak2024revisiting} have begun to address the need for real-world industrial data, highlighting the performance gap of existing methods when faced with such complexities.

\textbf{Human Gait\&Foot Datasets. } 
Gait analysis research has generated numerous datasets, often focusing on human identification or action recognition. CASIA Gait Dataset is a widely used early dataset that provides silhouette sequences of 124 subjects from 11 viewing angles~\cite{yu2006framework}. Similarly, the OU-ISIR Multi-View Large Population (OU-MVLP)~\cite{takemura2018multi} represent large-scale efforts, capturing over 10,000 subjects from 14 viewpoints, but typically rely on 2D representations like silhouettes or Gait Energy Images (GEIs)  derived from controlled captures. While effective for identification based on overall body shape and motion patterns, they lack detailed 3D surface information, especially of the distal segments like the feet. More recently, datasets such as Gait3D~\cite{zheng2022gait} have incorporated 3D information by providing SMPL model parameters recovered from videos captured in unconstrained environments. This allows for viewpoint and coarse shape analysis but still represents the body via a parametric model, not as dense, direct surface measurements suitable for detailed biomechanical analysis of specific joints or segments like the foot-ankle complex. 

Moreover, research focusing specifically on the human foot has led to the creation of datasets primarily capturing static 3D foot morphology. The Foot3D dataset~\cite{boyne2022find}, for instance, provides high-resolution 3D scans of feet in various static poses, collected using precise scanning equipment. Such datasets are invaluable for creating statistical shape models (e.g., the associated FIND model ) or training models for applications like custom orthotics fitting or virtual shoe try-on. Synthetic datasets like SynFoot~\cite{boyne2024found} have also been generated, often using a small number of real scans (from Foot3D) as a base, to provide large amounts of training data (e.g., images paired with normals) for tasks like image-based static foot reconstruction. However, the critical limitation of these foot-specific datasets is their static nature; they do not capture the dynamic changes in shape and relative segment orientation that occur during functional activities.

\textbf{Point Cloud Completion. }
The goal of point cloud completion is to infer a complete 3D shape from an incomplete input~\cite{fei2022comprehensive}. 
Early methods often employed 3D Convolutional Neural Networks (CNNs) on voxelized representations (e.g., GRNet~\cite{xie2020grnet}) or combined feature extraction with multiple folding-based decoders to generate complete point clouds (e.g., PCN~\cite{yuan2018pcn}). These methods sometimes struggled with capturing fine geometric details due to voxel resolution limits or limitations in modeling long-range dependencies. 
More recently, Transformers have shown great promise due to their ability to model global context and long-range interactions via self-attention mechanisms~\cite{fei2024progressive}. 
Architectures like PointTr~\cite{yu2021pointr}, AnchorFormer~\cite{chen2023anchorformer} apply transformer blocks directly to points or point patches, often achieving state-of-the-art results by better capturing structural relationships and generating more coherent shapes~\cite{fei2025point}. 
Generative approaches, including Generative Adversarial Networks (GANs), are also being explored. 
These methods aim to learn the underlying distribution of complete shapes and generate diverse and plausible completions, sometimes using hierarchical structures (e.g., SnowflakeNet~\cite{xiang2021snowflakenet}) to refine the output progressively~\cite{fei2025multi}.

%% file: Sections/Dataset_Description.tex
\section{Dataset}
\label{sec:dataset}

\subsection{Data Acquisition System}
\label{sec:data_acquisition_system}

The data for FootGait3D was collected using a custom-designed markerless motion capture system, previously described as the Point-cloud Foot Analysis (PFA) system~\cite{jiang2023fast}. This system was specifically engineered for fast and accurate evaluation of 3D foot-ankle kinematics during gait.

\begin{figure}[t]
    \centering
    \includegraphics[width=1\linewidth]{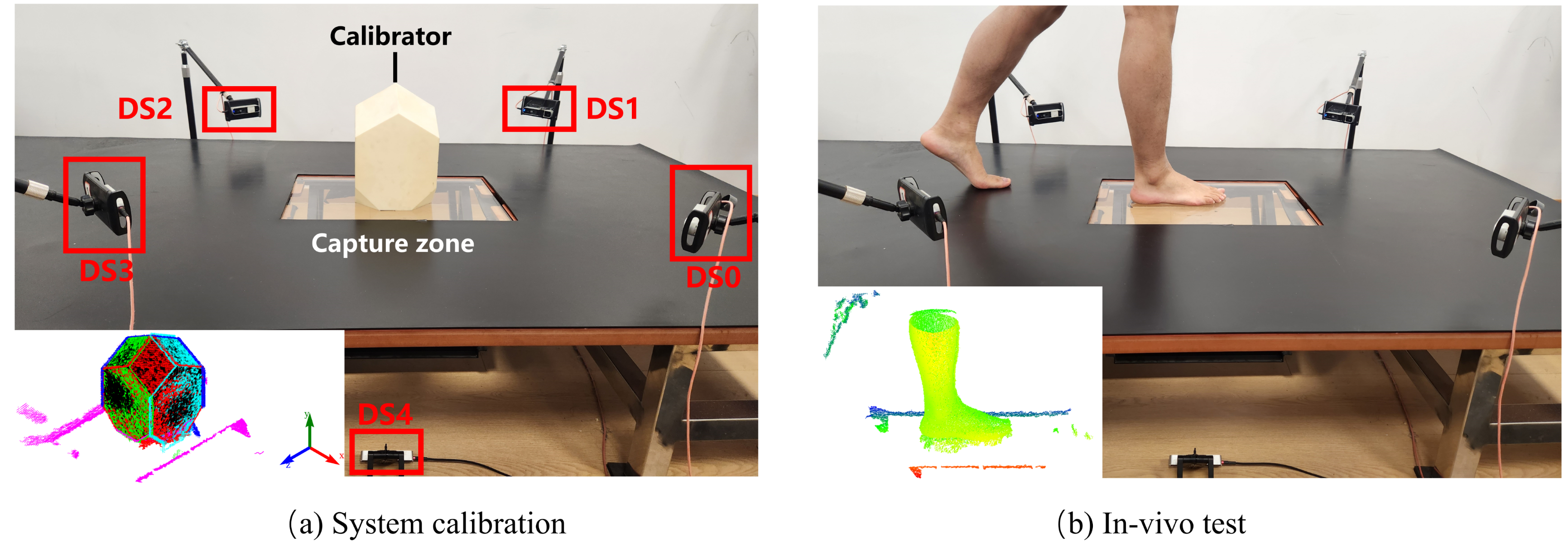}
    \vspace{-0.5cm}
    \caption{\textbf{Set up of the Point-cloud Foot Analysis (PFA) system.} DS0 to DS4 are five depth sensors. 
    (a) \textbf{System registration process.} In the center of the capture zone is a white calibrator, used for the system's coarse alignment process. The bottom-left corner shows the calibration object point cloud after fine registration using the pl-ICP algorithm. 
    (b) \textbf{Data acquisition process.} The bottom-left corner shows the raw foot point cloud before data preprocessing.}
\label{fig:PFA}
\vspace{-0.2cm}
\end{figure}

As shown in Figure~\ref{fig:PFA}, the core of the system consists of five Structure Core depth sensors (Occipital Inc., USA). 
Each sensor features a field of view (FoV) of $59^\circ$$\times$$46^\circ$$\times$$70^\circ$ and can achieve approximately 3mm precision within a 1-meter range at a capture rate of 30 FPS. 
To capture the foot-ankle complex surface comprehensively, four sensors (DS0-DS3) were positioned horizontally around a predefined capture zone, aimed at capturing the dorsal and side aspects of the foot.
Importantly, a fifth sensor (DS4) was placed beneath a transparent gait plate, mounted flush with the walkway surface, and tilted upwards to capture the plantar (bottom) surface of the foot during the stance phase while minimizing specular reflections. 
This multi-view setup ensures near-complete surface coverage of the foot–ankle complex during ground contact. 
Further, a global coordinate system was established with its origin at the geometric center of the transparent plate's upper surface, defining the X (anterior-posterior), Y (superior-inferior), and Z (medial-lateral) axes relative to the plate. 

Accurate spatial localization of each sensor is critical for merging the multi-view point clouds into a coherent representation. 
To achieve this, we implemented a robust coarse-to-fine registration procedure. 
Coarse registration involved using a custom-designed physical calibration object to determine the initial sensor alignment (see Figure~\ref{fig:PFA}a), while fine registration employed the point-to-plane Iterative Closest Point (pl-ICP) algorithm~\cite{chen1992object} to estimate the precise pose (transformation matrix relative to the global coordinate system) of each sensor.

\subsection{Recruitment \& Protocol}
A total of 46 subjects (mean age: 27.2±4.3 years, height: 174.5±5.7 cm, weight: 67.8±8.3 kg) participated in the data collection. 
All subjects all free from lower limb, foot, and ankle injuries or diseases. 
The study protocol was approved by the Institutional Review Board (IRB) of Fudan University, and informed consent was obtained from all participants prior to data collection. 
Participants were asked to walk barefoot at a self-selected comfortable speed back and forth along the walkway, ensuring their foot made full contact with the transparent plate within the designated capture zone (see Figure 1b). Each subject performed 8 gait tests after familiarization trials were conducted.

\subsection{Data Processing}
The raw multi-view depth maps captured by the system contained substantial background noise and often included the contralateral swing foot alongside the stance foot of interest, necessitating a comprehensive offline processing pipeline to generate clean, paired full-view and diverse partial-view point clouds of the stance foot.
Key stages include multi-view transformation and merging, stance phase isolation, noise removal and partial view synthesis.
\subsubsection{Preprocessing Pipeline}
\label{sec:proprocessing_pipeline}

\begin{figure}[t]
    \centering
    \includegraphics[width=1\linewidth]{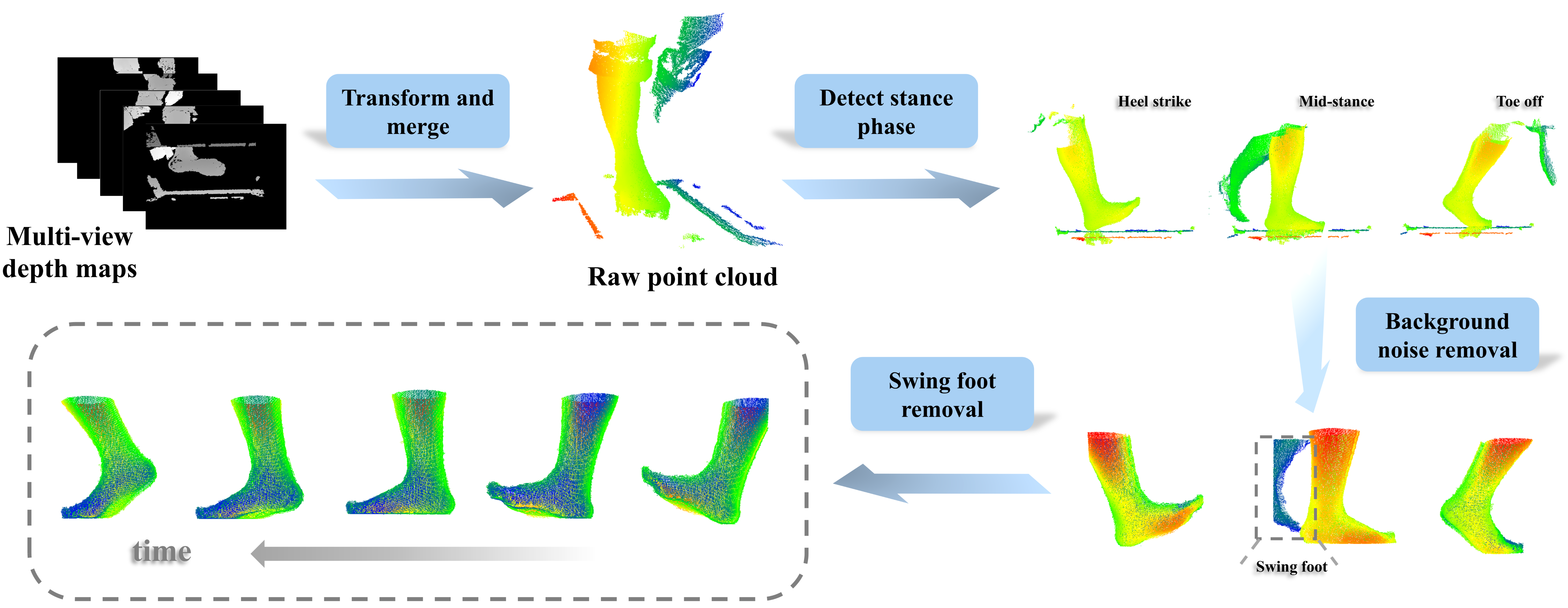}
    \vspace{-0.6cm}
    \caption{Data preprocessing pipeline. The core stages include multi-view transformation and fusion, stance phase isolation, noise reduction, and swing foot removal.}
\label{fig:preprocessing2}
\vspace{-0.2cm}
\end{figure}

The data preprocessing is illustrated in Figure~\ref{fig:preprocessing2}. 
In an offline processing pipeline, the saved depth maps from each camera were first converted to 3D point clouds (using the camera’s calibration to map depth pixels to 3D coordinates). 
Then, using the recorded timestamps and the pre-calculated sensor calibration matrices (from Sec.~\ref{sec:data_acquisition_system}), the point clouds from all five views corresponding to the same time instance were synchronized and transformed into the global coordinate system, then merged to form a single point cloud of the foot. 
Next, the merged point cloud was cropped using a predefined 3D bounding box corresponding to the capture zone to remove most background points (e.g., specular reflections). 

Subsequently, to generate the point cloud sequences corresponding to the stance phase, heel-strike (HS) and toe-off (TO) events were detected for each sequence. 
This was achieved by defining the ground surface as a virtual layer between two adjacent horizontal planes in the global coordinate system and counting the number of contacting foot points within this layer for each frame. 
HS and TO events were identified based on changes in this point count, ensuring all provided sequences represent the contact phase of gait. 
Finally, the contralateral swing foot was automatically removed by detecting the medial-lateral bounds between the feet. 
This step yields the clean Full-View (Ground Truth) point cloud for the stance foot during the stance phase, with each phase comprising approximately 28 to 36 frames.

\subsubsection{Partial Point Cloud Generation}
To create paired data (partial vs. complete) for completion benchmarks, partial point clouds were generated corresponding to scenarios with fewer cameras. 
This was achieved by selecting subsets of the calibrated sensor data before the merging step (from Sec.~\ref{sec:proprocessing_pipeline}). 
Importantly, the bottom camera (DS4) is always removed in these reduced-view scenarios, simulating a conventional non-transparent floor setup where the plantar surface is occluded. 
Further removal of the side views can simulate occlusions caused by the swing foot during the gait.
Accordingly, the view subsets are formed by different combinations of the remaining four cameras positioned around the foot:
\begin{itemize}[leftmargin=1em, itemsep=0pt, topsep=0pt, parsep=0pt]
    \item \textbf{4-View:} Merged data from DS0, DS1, DS2, DS3 (1 combination).
    \item \textbf{3-View:} Merged data from all possible combinations of 3 horizontal sensors (4 combinations: [DS0, DS1, DS2], [DS0, DS1, DS3], [DS0, DS2, DS3], [DS1, DS2, DS3])
    \item \textbf{2-View:} Merged data from all possible combinations of 2 horizontal sensors (6 combinations: [DS0, DS1], [DS0, DS2], [DS0, DS3], [DS1, DS2], [DS1, DS3], [DS2, DS3])
\end{itemize}
As the number of available camera views decreases, the extent of missing (occluded) regions in the foot point cloud progressively increases. 
To facilitate robustness evaluation under varying levels of view occlusion, our dataset includes graded versions ranging from complete five-view data to limited two-view data.

\subsubsection{Dataset Contents \& Quality}
\begin{figure}[t]
    \centering
    \includegraphics[width=1\linewidth]{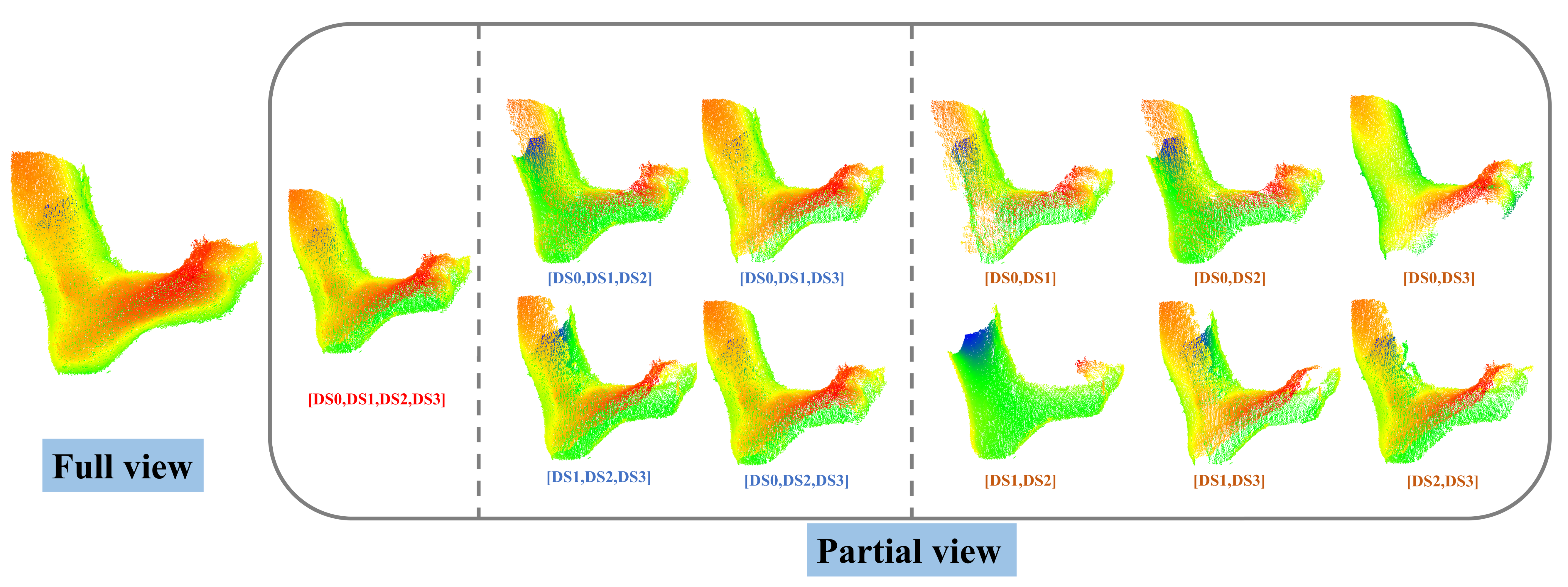}
    \vspace{-0.6cm}
    \caption{A paired complete and partial point clouds of an example instance in the FootGait3D dataset. From left to right: full 5-view reconstruction (DS0–DS4), 4-view, selected 3-view, and 2-view combinations.}
\label{fig:instance}
\vspace{-0.2cm}
\end{figure}

Each full-view ground truth point cloud was visually inspected before inclusion to ensure no significant surface regions were missing due to occlusions during capture. The final FootGait3D dataset comprises 8403 processed point clouds from 46 healthy subjects. 
Each instance includes the raw depth maps from all 5 sensors, the ground truth point cloud (8403 frames), the partial cloud from the 4 horizontal sensors (8403 frames), the four combinations of 3-view horizontal sensor data (33612 frames), and the six combinations of 2-view horizontal sensor data (50418 frames). In total, the dataset contains 100836 frames of point cloud data and 42015 depth maps (5 views $\times$ 8403 instances). 
Manual checks and cropping were also performed as needed to guarantee that each ground truth cloud is clean, representing only the stance foot without residual noise points (e.g., from the swing foot or background).
The acquisition system demonstrated high geometric fidelity with surface reconstruction accuracy below 2 mm root mean square error (RMSE). 
Figure~\ref{fig:instance} illustrates a paired complete and partial point cloud from a representative instance. 
The dataset reflects real-world dynamic capture conditions, including sensor noise and non-uniform sparsity, while the partial views offer realistic incompleteness patterns. 
Missing surface regions become more pronounced as fewer cameras are used. 
These graded occlusion levels provided reliable benchmarks for evaluating completion algorithms under practical constraints. 
Sensor calibration matrices are also provided alongside the data. 

%% file: Sections/Benchmark_and_Evaluations.tex
\section{Benchmark and Evaluations}
\label{sec:Benchmark}

\subsection{Benchmark Task}
The primary benchmark task defined for the FootGait3D dataset is point cloud completion. 
Given a partial point cloud representing an incomplete view of the foot-ankle complex during the gait contact phase, the goal is to predict the complete, dense point cloud corresponding to the ground truth for that same time instance.
This task challenges models to reconstruct detailed, dynamically changing geometry from various levels of realistic incompleteness caused by limited views. 
To quantitatively evaluate the performance of the point cloud completion methods, the following standard metrics were employed:
\begin{itemize}[leftmargin=1em, itemsep=0pt, topsep=0pt, parsep=0pt]
    \item \textbf{Chamfer Distance (CD):}  This metric measures the average closest point distance between predicted and ground truth point clouds~\cite{wu2021density}. CD-L1 (using L1 norm) and CD-L2 (using squared L2 norm) were both reported in this task. Lower values are better.The formal definitions of CD‑L1 and CD‑L2 for two point sets \(S_1\) and \(S_2\) are given in Eq.~\eqref{eq:cd}:

\begin{equation}
\begin{aligned}
\mathrm{CD}_{\text{L1}}(S_1,S_2) &=
\frac{1}{2|S_1|}\sum_{x\in S_1}\min_{y\in S_2}\lVert x-y\rVert_2
\;+\;
\frac{1}{2|S_2|}\sum_{y\in S_2}\min_{x\in S_1}\lVert y-x\rVert_2, \\[4pt]
\mathrm{CD}_{\text{L2}}(S_1,S_2) &=
\frac{1}{|S_1|}\sum_{x\in S_1}\min_{y\in S_2}\lVert x-y\rVert_2^{2}
\;+\;
\frac{1}{|S_2|}\sum_{y\in S_2}\min_{x\in S_1}\lVert y-x\rVert_2^{2}.
\end{aligned}
\tag{1}\label{eq:cd}
\end{equation}

Here \(S_1\) is the predicted point cloud and \(S_2\) is the ground‑truth point cloud; the terms \(\min_{y\in S_2} \lVert x - y \rVert_2\) and \(\min_{x\in S_1} \lVert y - x \rVert_2\) represent the Euclidean distance from each point to its nearest neighbour in the other set.

    \item \textbf{F1-Score:} This metric measures accuracy at a distance threshold $\tau$, balancing precision and recall. $\tau$ = 0.001 (1e-3) was used in this task. Higher values are better. The definition is given by:

\begin{equation}
\mathrm{F1}_\tau(S_1,S_2)=
\frac{%
      2
      \displaystyle\Bigl(\sum_{x\in S_1}\!\mathbf{1}\bigl[\min_{y\in S_2}\|x-y\|_2<\tau\bigr]\Bigr)\!
      \Bigl(\sum_{y\in S_2}\!\mathbf{1}\bigl[\min_{x\in S_1}\|y-x\|_2<\tau\bigr]\Bigr)
     }{%
      |S_1|\displaystyle\sum_{y\in S_2}\!\mathbf{1}\bigl[\min_{x\in S_1}\|y-x\|_2<\tau\bigr]+
      |S_2|\displaystyle\sum_{x\in S_1}\!\mathbf{1}\bigl[\min_{y\in S_2}\|x-y\|_2<\tau\bigr]
     },
\tag{2}\label{eq:f1}
\end{equation}

where \(S_1\) and \(S_2\) are the predicted and ground‑truth point‐cloud sets (with cardinalities \(|S_1|\) and \(|S_2|\)); \(\mathbf{1}[\cdot]\) is the indicator function.
\end{itemize}

\subsection{Baseline Methods}
To benchmark on FootGait3D, this work mainly evaluated representative point cloud completion methods in two categories: single-modal and multi-modal. 
The single-modal methods take only the partial point cloud as input, while the multi-modal methods leverage additional depth maps for improved completion. Single-modal methods include: \textbf{PCN}~\cite{yuan2018pcn}, \textbf{PoinTr}~\cite{yu2021pointr}, \textbf{AnchorFormer}~\cite{chen2023anchorformer}, \textbf{SnowflakeNet}~\cite{xiang2021snowflakenet}, and \textbf{PointAttN}~\cite{wang2024pointattn}.
Multi-modal methods include: 
\textbf{SVDFormer}~\cite{zhu2023svdformer}, \textbf{PointSea}~\cite{zhu2025pointsea}, \textbf{CSDN}~\cite{zhu2023csdn}, \textbf{EGIINet}~\cite{xu2024explicitly}, \textbf{XMFNet}~\cite{aiello2022cross}, and \textbf{MAENet}~\cite{liu2025maenet}.

\input{Tables/V2_Completion_performance_single}
\input{Tables/Completion_performance_multi}

\subsection{Experimental Setup}
All experiments were conducted on two hardware systems: one configured with three NVIDIA RTX3090 GPUs and the other with two NVIDIA RTX4090 GPUs. 
The dataset, consisting of 8403 total instances (frames), was split for training, validation, and testing based on the instance count: 5881 instances were allocated for training, 841 for validation, and 1681 for testing. 
Consequently, the total number of training, validation, or testing samples for each view condition is the base number multiplied by the number of combinations for that condition (e.g., 4 for 3-view). Specifically, there are 5881 (4-view), 23524 (3-view), and 35286 (2-view) samples in the training set; 841, 3364, and 5046 in the validation set; and 1681, 6724, and 10086 in the test set, respectively.

For all benchmark experiments, the input partial point clouds were uniformly downsampled to 2048 points. 
To promote robustness to varying degrees of incompleteness, each baseline model was trained uniformly on a combined set of partial-view inputs from the training data, including the single 4-view setting, all four 3-view settings, and all six 2-view settings. 
All models were trained using their official default hyperparameters.
To ensure a fair comparison while preserving the integrity of each model’s original design, all models were trained using their official default channels and output configurations. 
Specifically, all single-modal methods (PCN, PointTr, AnchorFormer, SnowflakeNet, and PointAttn) as well as the multi-modal SVDFormer and PointSea were trained to output 16384 points. 
For these models, the ground truth with 16384 points was used for loss computation. 
The remaining four multi-modal methods (CSDN, EGIINet, XMFNet, and MAENet) produce a default output of 2048 points and were therefore trained to predict point clouds at this resolution. 
Accordingly, the ground-truth point clouds were downsampled to 2048 points during training.

For multi-modal methods, the architectures were adapted to accommodate real-world multi-view inputs, as our dataset provides five-view depth maps along with corresponding camera extrinsic parameters.
For instance, in the original implementation of SVDFormer, three virtual depth images are rendered from fixed viewpoints to simulate self-views. In our setup, these synthetic views are replaced with actual five-view depth maps, and the real extrinsic parameters are incorporated into the feature fusion module to enable accurate spatial encoding. 
During evaluation, to mitigate the influence of output resolution differences on results, all 16384-point predictions and their corresponding ground-truth point clouds were uniformly downsampled to 2048 points using Farthest Point Sampling (FPS) prior to metric computation.

\subsection{Evaluations}
\label{sec:evaluation}
The quantitative results of baseline methods are summarized in Table~\ref{tab:footgait3d_single_V2} and Table~\ref{tab:footgait3d_multi}, presenting the comprehensive performance of both single-modal and multi-modal baseline methods on the FootGait3D benchmark.

\begin{figure}[t]
    \centering
    \includegraphics[width=1\linewidth]{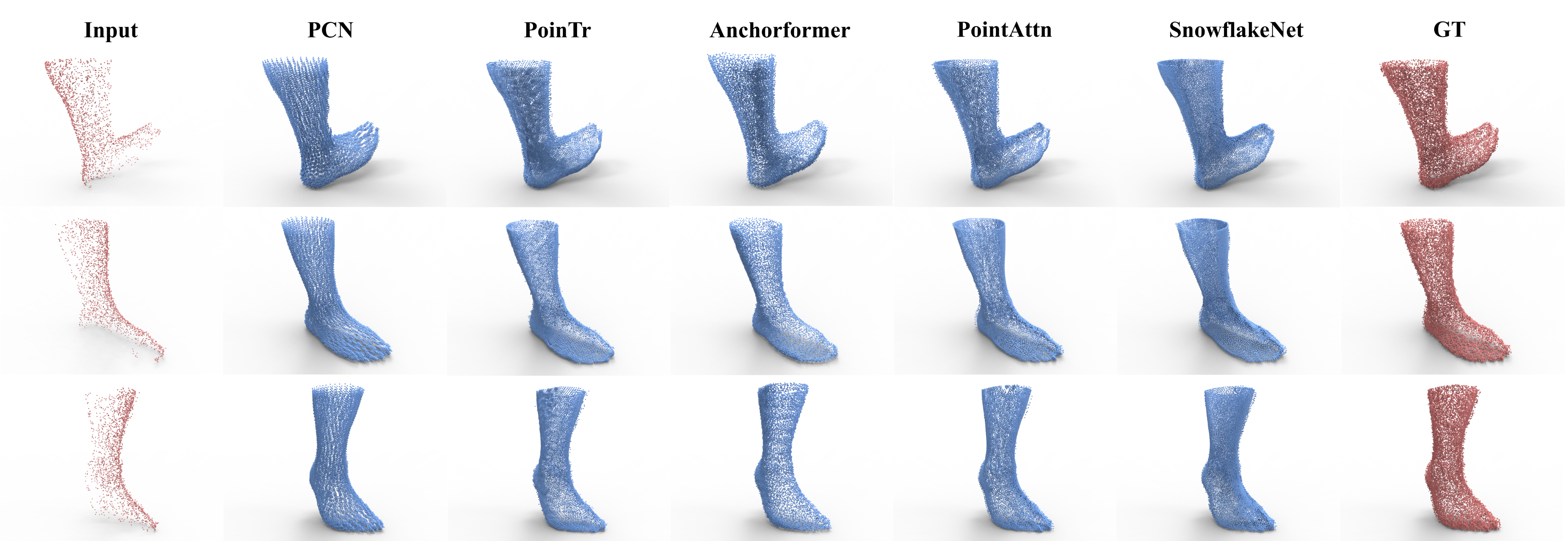}
    \vspace{-0.7cm}
    \caption{Visual comparison of \textbf{single-modal} point cloud completion results on the FootGait3D under the challenging 2-view[DS1, DS2] input condition. From top to bottom, each row corresponds to a partial input captured at different gait events: heel strike, mid-stance, and toe off. All predictions consist of 16384 points.}
\label{fig:2view-single}
\end{figure}

Single-modal methods, which only rely on partial point cloud inputs, generally struggle to handle the noise, non-uniform sparsity, and complex deformations present in the FootGait3D dataset (Figure~\ref{fig:2view-single}). 
Among these methods, SnowflakeNet and PointAttn achieve the best results. 
The success of SnowflakeNet can be attributed to its hierarchical structure, which progressively refined the output and enabled the generation of fine geometric details. 
Similarly, PointAttn effectively leverages self-attention mechanisms to model global context and long-range interactions, resulting in high-fidelity shape completion. 
In contrast, earlier methods such as PCN rely on simpler folding-based decoders and therefore struggle to capture intricate details and structural relationships. 
As a result, they exhibit inferior performance on this challenging real-world dataset.
Interestingly, different models react differently to the input data. 
For instance, PointAttn performs better under the 3-view and 2-view conditions than under the 4-view condition, suggesting that its architecture gained greater robustness from the larger quantity and variety of sparse training samples. 
Conversely, AnchorFormer’s performance steadily degrades as the number of views decreases, indicating its greater sensitivity to the density and completeness of the input data.


\begin{figure}[t]
    \centering
    \includegraphics[width=1\linewidth]{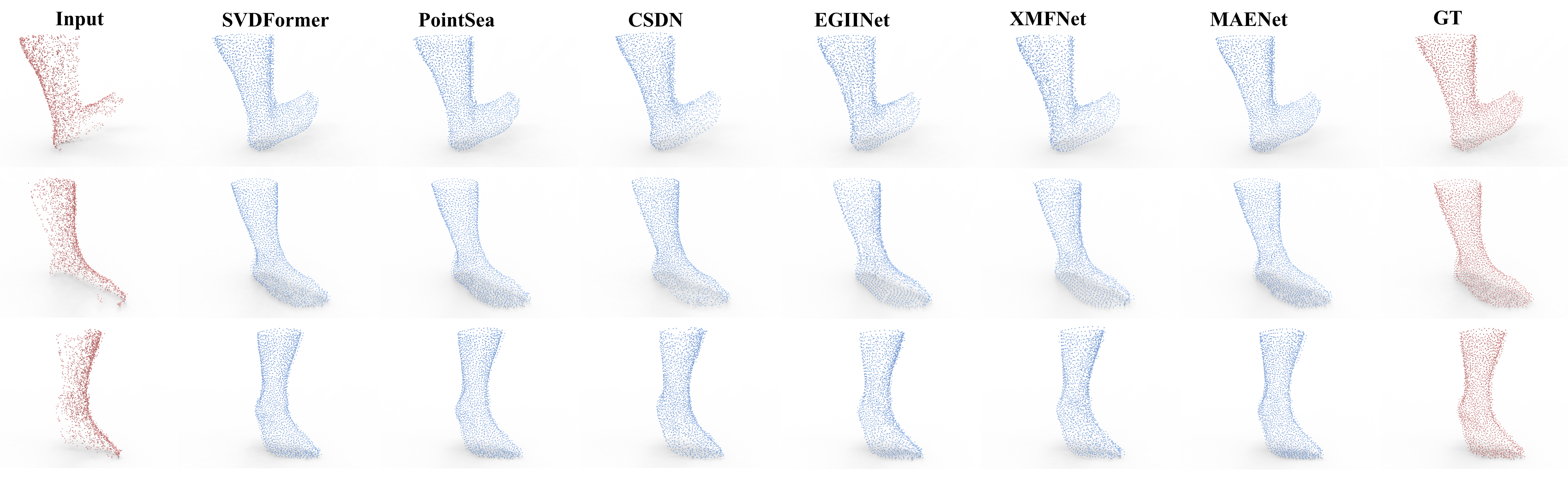}
    \vspace{-0.8cm}
    \caption{Visual comparison of \textbf{multi-modal} point cloud completion results on the FootGait3D under the challenging 2-view[DS1, DS2] input condition. From top to bottom, each row corresponds to a partial input captured at different gait events: heel strike, mid-stance, and toe off. All predictions consist of 2048 output points.}
\label{fig:2view-multi}
\vspace{-0.2cm}
\end{figure}

The multi-modal methods, adapted to utilize the real multi-view depth information from our FootGait3D dataset, generally demonstrate the benefit of fusing information from multiple perspectives.
These methods show high stability and robustness across all view conditions, since they perform well across the 4-view, 3-view, and 2-view settings.
As illustrated in Figure~\ref{fig:2view-multi}, all models successfully reconstruct plausible foot shapes even from challenging 2-view inputs, effectively capturing the geometry. 
It is worth noting that PointSea and SVDFormer generated high-resolution outputs of 16384 points. 
At this high resolution, both of them achieve superior CD values and F1-score, indicating highly accurate dense reconstructions. 
Even when their outputs are downsampled to 2048 points for fair comparison, they still maintain the best performance among the multi-modal methods. 
These results demonstrate that their architectures are particularly effective at integrating multi-view information to capture the underlying geometric structure for real-world reconstruction.


%% file: Tables/V2_Completion_performance_single.tex



\begin{table}[t]
  \centering
   \caption{Completion performance on \textsc{FootGait3D} \textbf{(single-modal models)}. 
           Lower CD-L1/CD-L2 ($\!\times10^{3}$) are better ($\downarrow$); 
           higher F1-score at 1\textperthousand\ of the bounding-box diagonal is better ($\uparrow$). 
           Values in parentheses correspond to results when the predicted point number is 16384.}
  \label{tab:footgait3d_single_V2}
  \resizebox{\linewidth}{!}{%
  \begin{tabular}{l ccc ccc ccc}
    \toprule
    \multirow{2}{*}{\textbf{Model}} &
    \multicolumn{3}{c}{\textbf{4-View}} &
    \multicolumn{3}{c}{\textbf{3-View}} &
    \multicolumn{3}{c}{\textbf{2-View}} \\
    \cmidrule(lr){2-4} \cmidrule(lr){5-7} \cmidrule(lr){8-10}
      & CD-L1$\downarrow$ & CD-L2$\downarrow$ & F1$\uparrow$
      & CD-L1$\downarrow$ & CD-L2$\downarrow$ & F1$\uparrow$
      & CD-L1$\downarrow$ & CD-L2$\downarrow$ & F1$\uparrow$ \\
    \midrule
    \multirow{1}{*}{PCN~\cite{yuan2018pcn}} 
      & \makecell{19.278 \\ (10.240)} & \makecell{0.855 \\ (0.273)} & \makecell{0.952 \\ (0.933)} 
      & \makecell{19.207 \\ (10.321)} & \makecell{0.860 \\ (0.288)} & \makecell{0.951 \\ (0.991)} 
      & \makecell{19.688 \\ (10.801)} & \makecell{0.907 \\ (0.320)} & \makecell{0.941 \\ (0.988)} \\
    \multirow{1}{*}{PoinTr~\cite{yu2021pointr}} 
      & \makecell{18.287 \\ (9.242)} & \makecell{0.782 \\ (0.229)} & \makecell{0.960 \\ (0.994)}
      & \makecell{17.828 \\ (8.679)} & \makecell{0.743 \\ (0.201)} & \makecell{0.968 \\ (0.995)}
      & \makecell{17.843 \\ (8.598)} & \makecell{0.860 \\ (0.221)} & \makecell{0.969 \\ (0.995)} \\
    \multirow{1}{*}{Anchorformer~\cite{chen2023anchorformer}} 
      & \makecell{18.944 \\ (8.767)} & \makecell{0.417 \\ (0.208)} & \makecell{0.952 \\ (0.996)} 
      & \makecell{18.174 \\ (9.268)} & \makecell{0.389 \\ (0.236)} & \makecell{0.965 \\ (0.994)} 
      & \makecell{20.880 \\ (11.930)} & \makecell{0.514 \\ (0.193)} & \makecell{0.911 \\ (0.981)} \\
    \multirow{1}{*}{SnowflakeNet~\cite{xiang2021snowflakenet}}
      & \makecell{17.671\\ (8.430)} & \makecell{0.720 \\ (0.181)} & \makecell{0.975 \\ (0.997)} 
      & \makecell{17.720 \\ (8.486)} & \makecell{0.727 \\ (0.186)} & \makecell{0.973 \\ (0.997)} 
      & \makecell{17.966 \\ (8.825)} & \makecell{0.810\\ (0.212)} & \makecell{0.969 \\ (0.996)} \\
    \multirow{1}{*}{PointAttn~\cite{wang2024pointattn}} 
      & \makecell{17.438 \\ (8.365)} & \makecell{0.703 \\ (0.179)} & \makecell{0.976 \\ (0.997)} 
      & \makecell{17.351 \\ (8.190)} & \makecell{0.697 \\ (0.173)} & \makecell{0.978 \\ (0.997} 
      & \makecell{17.477 \\ (8.354)} & \makecell{0.711 \\ (0.183)} & \makecell{0.976 \\ (0.997)} \\
    \bottomrule
  \end{tabular}
  }
\vspace{-0.2cm}
\end{table}


%% file: Tables/Completion_performance_multi.tex



\begin{table}[t]
  \centering
  \caption{Completion performance on \textsc{FootGait3D} \textbf{(multi-modal models)}. 
           Lower CD-L1/CD-L2 ($\!\times10^{3}$) are better ($\downarrow$); 
           higher F1-score at 1\textperthousand\ of the bounding-box diagonal is better ($\uparrow$). Values in parentheses correspond to results when the predicted point number is 16384.}
  \label{tab:footgait3d_multi}
  \resizebox{\linewidth}{!}{%
  \begin{tabular}{lccccccccc}
    \toprule
    \multirow{2}{*}{\textbf{Model}} &
      \multicolumn{3}{c}{\textbf{4-View}} &
      \multicolumn{3}{c}{\textbf{3-View}} &
      \multicolumn{3}{c}{\textbf{2-View}} \\
    \cmidrule(lr){2-4}\cmidrule(lr){5-7}\cmidrule(lr){8-10}
      & CD-L1$\downarrow$ & CD-L2$\downarrow$ & F1$\uparrow$
      & CD-L1$\downarrow$ & CD-L2$\downarrow$ & F1$\uparrow$
      & CD-L1$\downarrow$ & CD-L2$\downarrow$ & F1$\uparrow$ \\
    \midrule
    \multirow{1}{*}{SVDFormer~\cite{zhu2023svdformer}}
     & \makecell{17.299 \\ (8.143)} & \makecell{0.690 \\ (0.169)} & \makecell{0.978 \\ (0.998)} 
     & \makecell{17.304 \\ (8.139)} & \makecell{0.692 \\ (0.169)} & \makecell{0.979 \\ (0.998)} 
     & \makecell{17.410 \\ (8.292)} & \makecell{0.702 \\ (0.177)} & \makecell{0.978 \\ (0.997)} \\
    \multirow{1}{*}{PointSea~\cite{zhu2025pointsea}}  
    & \makecell{17.187 \\ (7.987)} & \makecell{0.681 \\ (0.162)} & \makecell{0.980 \\ (0.998)} 
    & \makecell{17.215 \\ (7.980)} & \makecell{0.685 \\ (0.162)} & \makecell{0.980 \\ (0.998)} 
    & \makecell{17.293 \\ (8.118)} & \makecell{0.692 \\ (0.169)} & \makecell{0.979\\ (0.998)} \\
    CSDN~\cite{zhu2023csdn}                    &19.358 & 0.881 & 0.936 &  19.247 & 0.885 & 0.949 &19.639 & 0.932& 0.932\\
    EGIINet ~\cite{xu2024explicitly}                &18.290 &0.789 & 0.955&17.932 & 0.759&0.962 &18.724 & 0.956& 0.947\\
    XMFNet~\cite{aiello2022cross}   & 18.858&0.850 &0.936& 18.423 & 0.806& 0.949 &18.416 &0.924  & 0.955\\
    MAENet~\cite{liu2025maenet}       & 17.672& 0.728& 0.970& 17.724& 0.735& 0.968 & 17.937& 0.756 &0.965 \\
    \bottomrule
  \end{tabular}
  }
\vspace{-0.2cm}
\end{table}

%% file: Sections/Discussion.tex
\section{Discussion}
\label{sec:discussion}
By providing dynamic and anatomically detailed point cloud data of the foot–ankle complex with paired partial and complete samples, FootGait3D aims to drive progress in biomechanics research and computer vision benchmarks.
In this section, we explore the broader impact of our dataset, address its current limitations, and propose avenues for future investigation.

\textbf{Insights on FootGait3D Benchmark. }
The benchmark evaluations presented in Sec.~\ref{sec:evaluation} reveal several key findings. 
(I) First, state-of-the-art point cloud completion methods, when trained and tested on FootGait3D, generally perform worse than on cleaner, synthetic datasets such as PCN or ShapeNet. 
This performance gap underscores the substantial challenges posed by real-world dynamic data, which is often affected by noise, non-uniform sparsity, and complex deformable motion. 
The degradation in performance, particularly in methods that perform well on synthetic benchmarks, highlights the importance of datasets like FootGait3D in evaluating models under realistic and challenging conditions.
(II) Moreover, the point cloud completion performance does not consistently decline as the number of input views decreases. 
The performance of single-modal methods is particularly varied, which is influenced by the trade-off between the density of an individual input and the total diversity of the training data. 
Moreover, the multi-modal methods demonstrate remarkable robustness, maintaining highly stable performance across all view conditions. 
This graded benchmark is therefore essential, as it not only quantitatively captures the increasing challenge posed by reduced spatial coverage but also enables a systematic assessment of the robustness and specific sensitivities of point cloud completion methods under varying degrees of data sparsity and occlusion.

\textbf{Limitations and Future Work. }
A primary limitation of the FootGait3D dataset is the imperfect synchronization among the five depth cameras used during data acquisition. 
Although a software-based synchronization mechanism was implemented to temporally align the devices, subtle inter-camera timing offsets remained. 
These desynchronization effects are particularly obvious during rapid foot movements occurring at the initial and terminal phases of the gait cycle—such as heel-strike and toe-off—where even slight temporal misalignments can lead to inconsistent or distorted multi-view point clouds, ultimately compromising reconstruction accuracy. 
To mitigate this issue and ensure data quality, sequence segments with noticeable misalignment were manually filtered.
However, this filtering procedure introduces another limitation: some gait cycles in the final released dataset are incomplete, with missing frames at the start or end of the cycle.

To address this limitation, future work will focus on hardware upgrades and improved timing control. 
We plan to adopt higher-frame-rate depth sensors and implement more robust synchronization methods (potentially hardware-triggered capture) to more tightly align the multi-camera data streams. 
These improvements are expected to enable the capture of more complete and reliable gait sequences, even during fast movements, thereby further enhancing the fidelity of FootGait3D.

%% file: appendix.tex





\section*{Supplementary Material}
\renewcommand{\thesubsection}{\Alph{subsection}} 
In this supplementary material, we provide additional information and extended results to complement our main paper. 
The primary aim of this supplementary material is to provide a comprehensive overview of the dataset characteristics and to present extended results for the benchmarked point cloud completion methods. 

\subsection{Dataset characteristics}
The size of the input point clouds varies significantly depending on the number of camera views and their specific combinations. Analyzing this distribution is essential for interpreting model performance, as point cloud density could affect both feature extraction and the completion process. 
Figure 6 illustrates the distribution of the number of points for the Ground Truth (GT, 5-view), 4-view, 3-view, and 2-view conditions using violin plots overlaid with box plots.

\begin{figure}[t]
    \centering
    \includegraphics[width=1\linewidth]{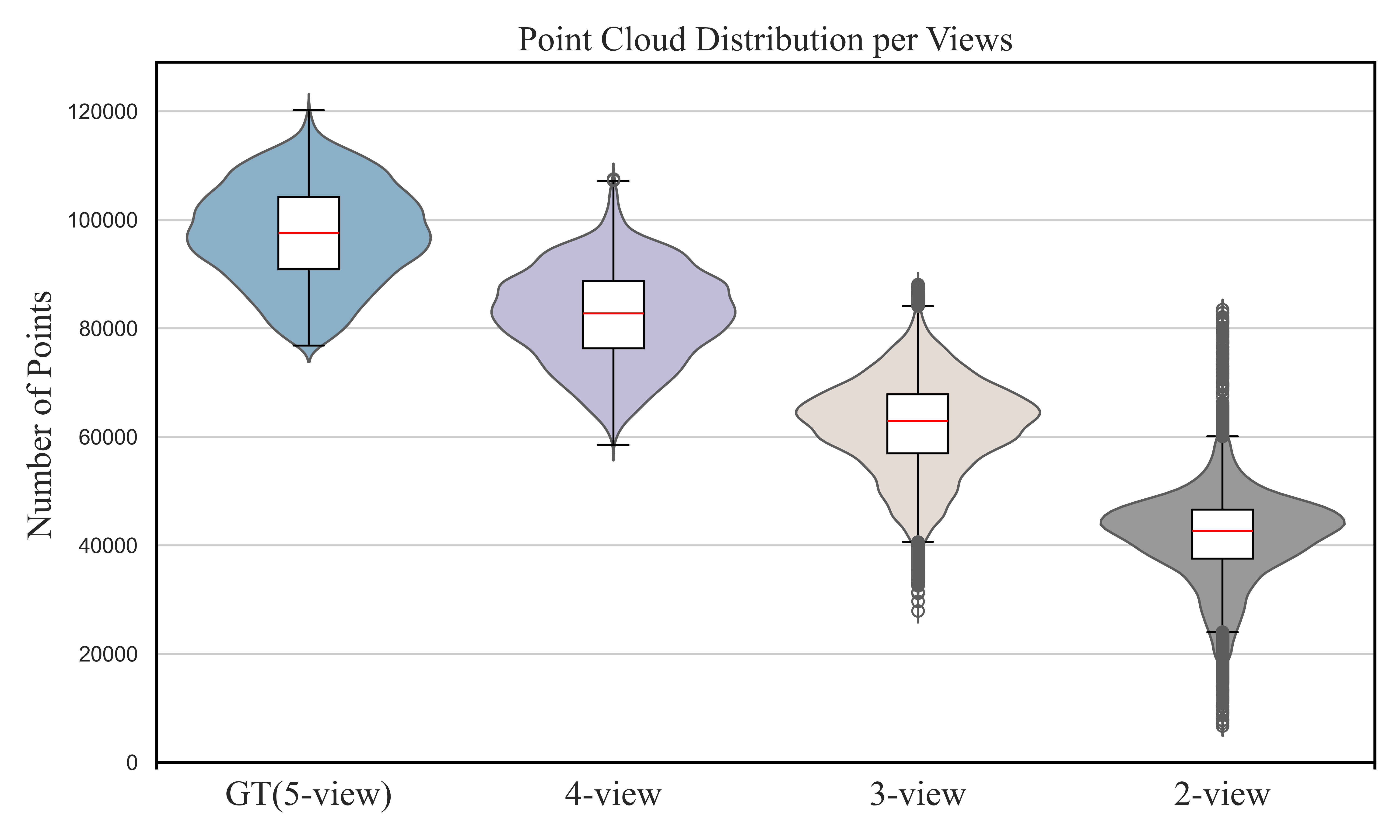}
    \vspace{-0.8cm}
    \caption{Point cloud density distribution across different view conditions in the FootGait3D dataset. The violin plots showed the probability density of the data at different point numbers, while the embedded box plots indicate the median, quartiles, and potential outliers.}
\label{fig:dataset_distribution}
\end{figure}

As depicted in Figure~\ref{fig:dataset_distribution}, the maximum density for all view conditions (widest part of the violin) occurs around the median point count for that condition. 
The ground truth (5-view) and 4-view conditions exhibit relatively stable and high point count distributions. 
For the ground-truth, point clouds are primarily distributed between approximately 78000 and 120000 points.
The 4-view inputs, generated by excluding the bottom camera (DS4), show a slightly reduced but still dense distribution, with point counts mostly ranging from 60000 to 106000. 
The 3-view and 2-view conditions, as expected, contain progressively fewer points. 
The 3-view samples were primarily distributed between 40000 and 80000 points, whereas the 2-view samples largely range from 20000 to 60000 points.

Notably, both the 3-view and 2-view conditions exhibit a higher presence of outliers, as indicated by individual points extending beyond the whiskers of the box plots. 
This reflect increased variability in point counts when fewer cameras were active. 
In particular, the 2-view condition display a pronounced widening around the 42000-point mark in its violin plot, suggesting a strong concentration of samples near this value despite the wider range and the existence of sparser instances. 
Such variability in point density—especially under the more challenging 2-view and 3-view scenarios—further increases the complexity of the completion task.

\subsection{Additional Results}

To provide a more comprehensive understanding of the models' performance, this section presents additional results for point cloud completion under 3-view and 4-view input conditions.

\subsubsection{Single-Modal Methods}
\begin{figure}[t]
    \centering
    \includegraphics[width=1\linewidth]{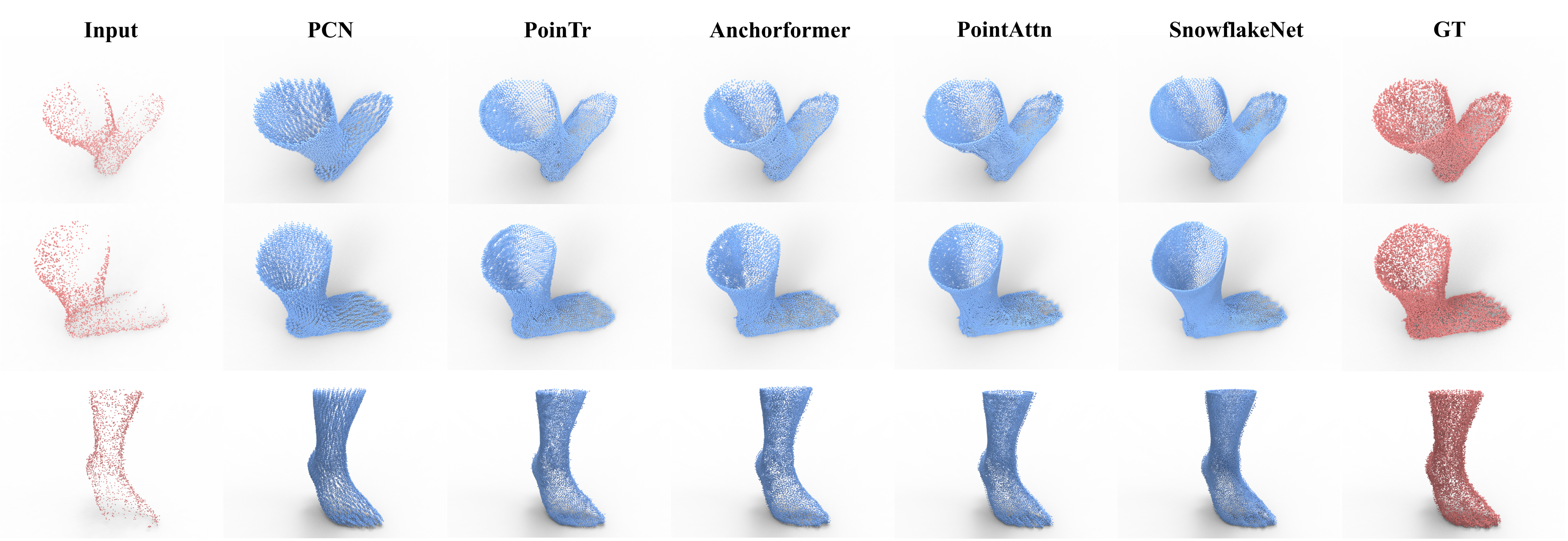}
    \vspace{-0.8cm}
    \caption{Visual comparison of \textbf{single-modal} point cloud completion results on the FootGait3D under the challenging 3-view[DS0, DS1, DS2] input condition. From top to bottom, each row corresponds to a partial input captured at different gait events: heel strike, mid-stance, and toe off. All predictions consist of 16384 points.}
\label{fig:3view-single}
\end{figure}

\begin{figure}[t]
    \centering
    \includegraphics[width=1\linewidth]{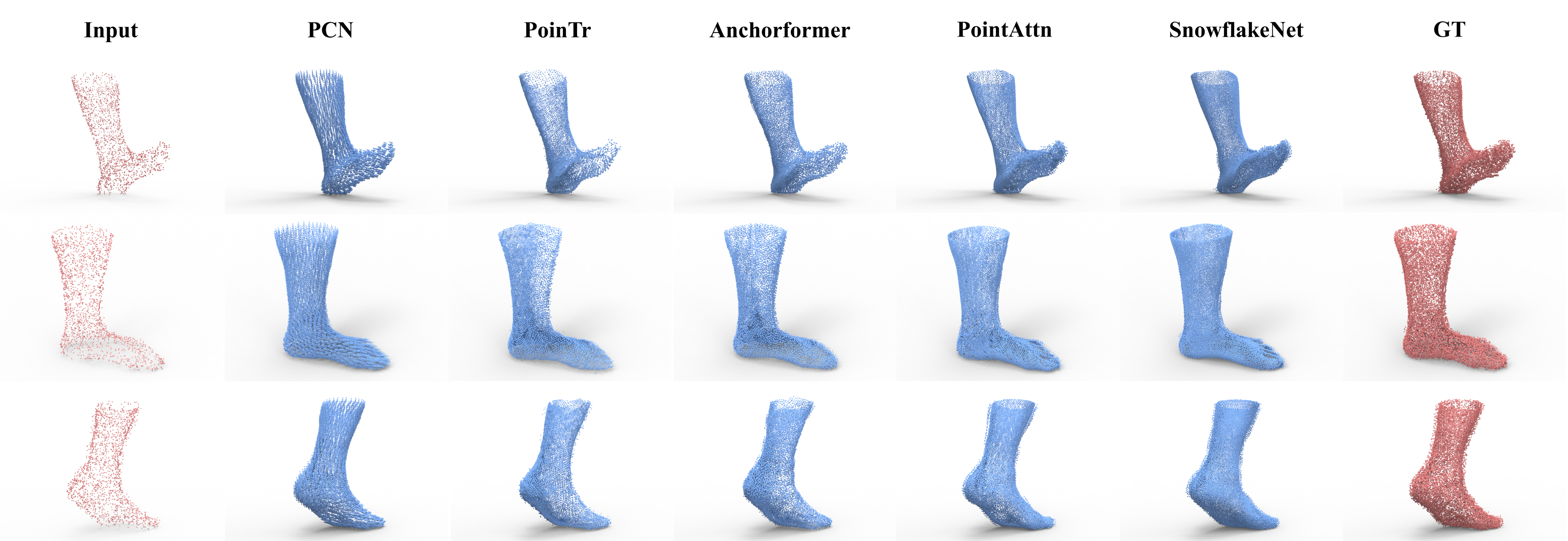}
    \vspace{-0.8cm}
    \caption{Visual comparison of \textbf{single-modal} point cloud completion results on the FootGait3D under the challenging 4-view[DS0, DS1, DS2, DS3] input condition. From top to bottom, each row corresponds to a partial input captured at different gait events: heel strike, mid-stance, and toe off. All predictions consist of 16384 points.}
\label{fig:4view-single}
\end{figure}

Figure~\ref{fig:3view-single} and Figure~\ref{fig:4view-single} illustrate the completion results of single-modal methods on representative samples from the FootGait3D test set under 3-view and 4-view input conditions, respectively. 
The input partial point clouds were shown alongside the predicted complete point clouds and the ground truth. 
All predictions consisted of 16384 points.

The performance trends of single-modal methods under the 3-view and 4-view conditions largely align with those reported for the 2-view scenario in the main paper. 
Notably, PCN~\cite{yuan2018pcn} consistently underperforms across all view conditions.
This suggests that its architecture with default configurations is likely ill-suited to the non-uniform sparsity and deformable structure characteristic of the FootGait3D data, regardless of the input point density.
An interesting observation pertains to PointTr~\cite{yu2021pointr}. 
While most models exhibit improved or stable performance with an increasing number of input views (i.e., denser partial point clouds), PointAttn performs worst under the 4-view condition—despite this setting providing the highest point density per sample. In contrast, its performance is comparatively better with 2-view and 3-view inputs.

\subsubsection{Multi-Modal Methods}
\begin{figure}[t]
    \centering
    \includegraphics[width=1\linewidth]{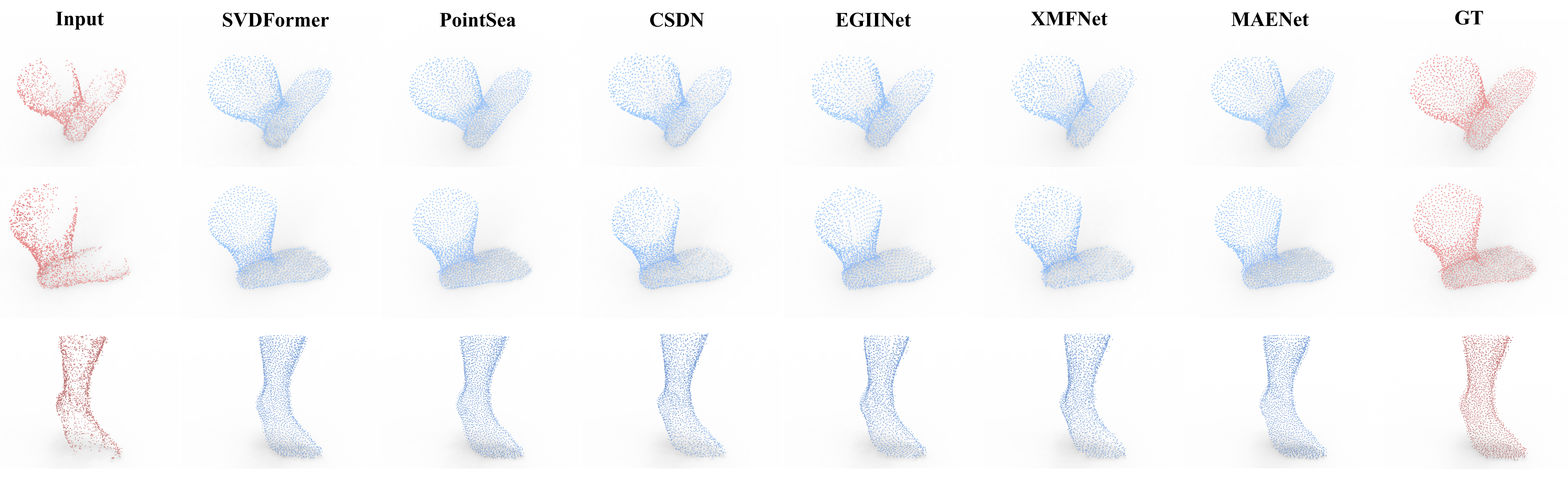}
    \vspace{-0.8cm}
    \caption{Visual comparison of \textbf{multi-modal} point cloud completion results on the FootGait3D under the challenging 3-view[DS0, DS1, DS2] input condition. From top to bottom, each row corresponds to a partial input captured at different gait events: heel strike, mid-stance, and toe off. All predictions consist of 2048 points.}
\label{fig:3view-multi}
\end{figure}

\begin{figure}[t]
    \centering
    \includegraphics[width=1\linewidth]{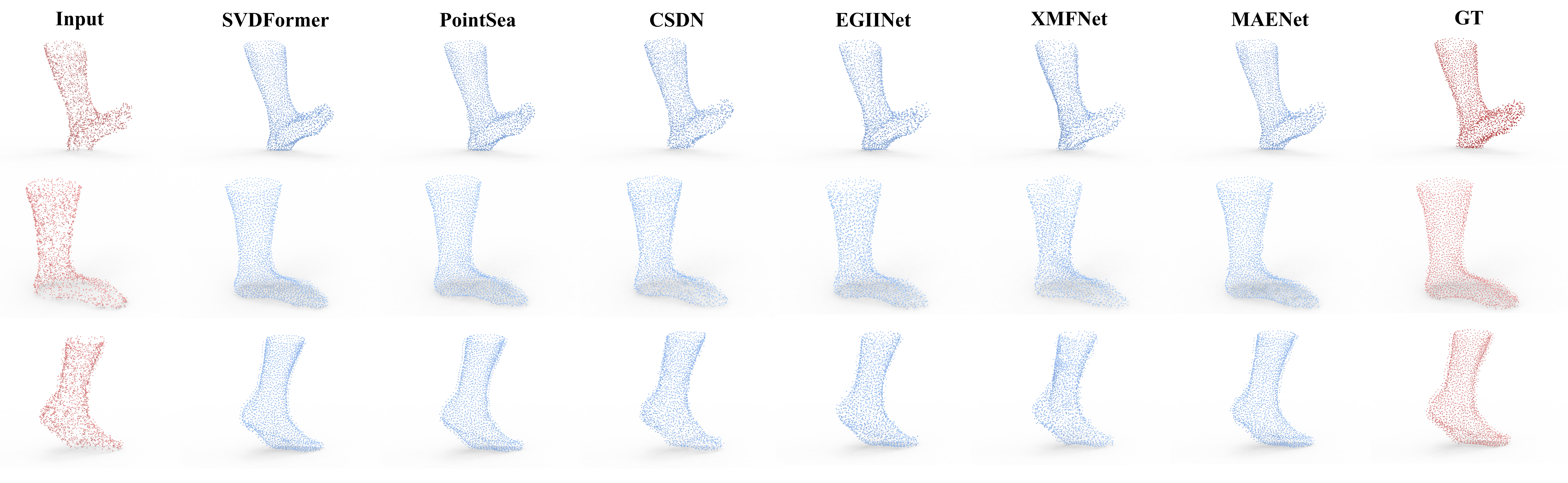}
    \vspace{-0.8cm}
    \caption{Visual comparison of \textbf{multi-modal} point cloud completion results on the FootGait3D under the challenging 4-view[DS0, DS1, DS2, DS3] input condition. From top to bottom, each row corresponds to a partial input captured at different gait events: heel strike, mid-stance, and toe off. All predictions consist of 2048 points.}
\label{fig:4view-multi}
\end{figure}

Figure~\ref{fig:3view-multi} and Figure~\ref{fig:4view-multi} illustrate the completion results of multi-modal methods on representative samples from the FootGait3D test set under 3-view and 4-view input conditions, respectively. 
The input partial point clouds are shown alongside the predicted complete point clouds and the ground truth. 
All predictions consist of 2048 points.
The multi-modal methods consistently achieve robust performance in reconstructing the overall foot shape across both view conditions, further validating the findings observed in the 2-view scenario. 
This suggests that the additional depth information from multiple views could lead to a more robust completion process across different view conditions. 
When reconstructing fine anatomical details such as the toe region, the performance gaps among different multi-modal models are more obvious under different view conditions.
SVDFormer~\cite{zhu2023svdformer} and PointSea~\cite{zhu2025pointsea} achieve the highest fidelity in reconstructing both the overall shape and fine-grained details. 
While all methods demonstrate robust global shape reconstruction, variations in the accuracy of subtle local features provide a more discriminative basis for comparison.
